\documentclass{article}
\usepackage{spconf,amsmath,graphicx, amssymb}
\usepackage{multicol, multirow}
\usepackage{soul}
\usepackage[shortlabels]{enumitem}
\setlist{nosep, leftmargin=14pt}
\usepackage{mwe} 
\usepackage{subcaption}
\usepackage{pgfplots}

\let\vec\mathbf

\title{End-to-end Deformable Attention Graph Neural Network for Single-view Liver Mesh Reconstruction}

\name{Matej Gazda $^{\star}$ $^{\dagger}$ \qquad Peter Drotar $^{\dagger}$\qquad Liset Vázquez Romaguera $^{\star}$ \qquad Samuel Kadoury \sthanks{Matej Gazda performed the work while at Ecole Polytechnique de Montréal, Montréal, QC H3C 3A7, Canada}}

\address{$^{\star}$ Polytechnique Montréal, Montréal, QC H3C 3A7, Canada \\ $^{\dagger}$ Intelligent Information Systems Laboratory, Technical University of Kosice, Kosice 040 12, Slovakia}

\begin{document}
%
\maketitle
\begin{abstract}
Intensity modulated radiotherapy (IMRT) is one of the most common modalities for treating cancer patients. One of the biggest challenges is precise treatment delivery that accounts for varying motion patterns originating from free-breathing. Currently, image-guided solutions for IMRT is limited to 2D guidance due to the complexity of 3D tracking solutions. 
We propose a novel end-to-end attention graph neural network model that generates in real-time a triangular shape of the liver based on a reference segmentation obtained at the preoperative phase and a 2D MRI coronal slice taken during the treatment. Graph neural networks work directly with graph data and can capture hidden patterns in non-Euclidean domains. Furthermore, contrary to existing methods, it produces the shape entirely in a mesh structure and correctly infers mesh shape and position based on a surrogate image. We define two on-the-fly approaches to make the correspondence of liver mesh vertices with 2D images obtained during treatment. Furthermore, we introduce a novel task-specific identity loss to constrain the deformation of the liver in the graph neural network to limit phenomenons such as flying vertices or mesh holes. The proposed method achieves results with an average error of $3.06 \pm 0.7$ mm and Chamfer distance with L2 norm of $63.14 \pm 27.28$.
\end{abstract}
\section{Introduction}
\label{sec:intro}

One of the most commonly used radiotherapy treatment is intensity modulated radiotherapy (IMRT), which consists in the delivery of tightly targeted radiation beams from outside the body. However, it faces complex challenges when experiencing important motion displacements, hence posing a great risk of dose administration to healthy tissue, such as in the liver \cite{stera2021liver}. Consequently, respiratory motion compensation is an important part of radiotherapy and other non-invasive interventions \cite{lu2018respiratory}. To avoid unnecessary damage due to organ displacement caused by respiratory motion, the treated organ must be located and imaged at all times. Unfortunately, image acquisition during treatment with IMRT is limited to 2D cine slices due to time complexity, therefore resulting in a lack of information in out-of-plane data for tumor targeting. Real-time 3D motion tracking of organs would provide necessary tools to accurately follow tumor targets.

Several methods, based on convolutional neural networks (CNNs) have been proposed to tackle the problem of modeling 3D data based on 2D signals. Mezheritsky et al. \cite{mezheritsky2022population} proposed a method that warps the reference volume with the output of a convolutional autoencoder, thus recovering 3D deformation fields with only a pre-treatment volume and a single live 2D image. Cerrolaza et al. \cite{cerrolaza20183d} proposed a 3D ultrasound fetal skull reconstruction method based on standard 2D ultrasound views of the head using a reconstructive conditional variational autoencoder. Girdhar et al. \cite{girdhar2016learning} investigated tackling a number of tasks including voxel prediction from 2D images and 3D model retrieval. However, methods based on CNNs achieved partial success, since they rely on fixed-size inputs. The volumes might have different sizes across scans, body types, and/or machines. The requirement of having volumes reshaped might result in information loss. 

Recent advances in graph neural networks sparked new advances in many domains, including in medical imaging \cite{ahmedt2021graph}. Lu et al. \cite{lu2021dynamic} leveraged dynamic spatio-temporal graph neural network for cardiac motion analysis. Graph neural networks operate on graph objects, which contrasts to representations obtained by CNN that might lose important surface details. The mesh has more desirable properties for many applications, because they are lightweight, have more shape modeling details, and are better suited for simulating deformations \cite{ahmedt2021graph}.


\begin{figure}[h]
    \centering
    \includegraphics[width=0.5\textwidth]{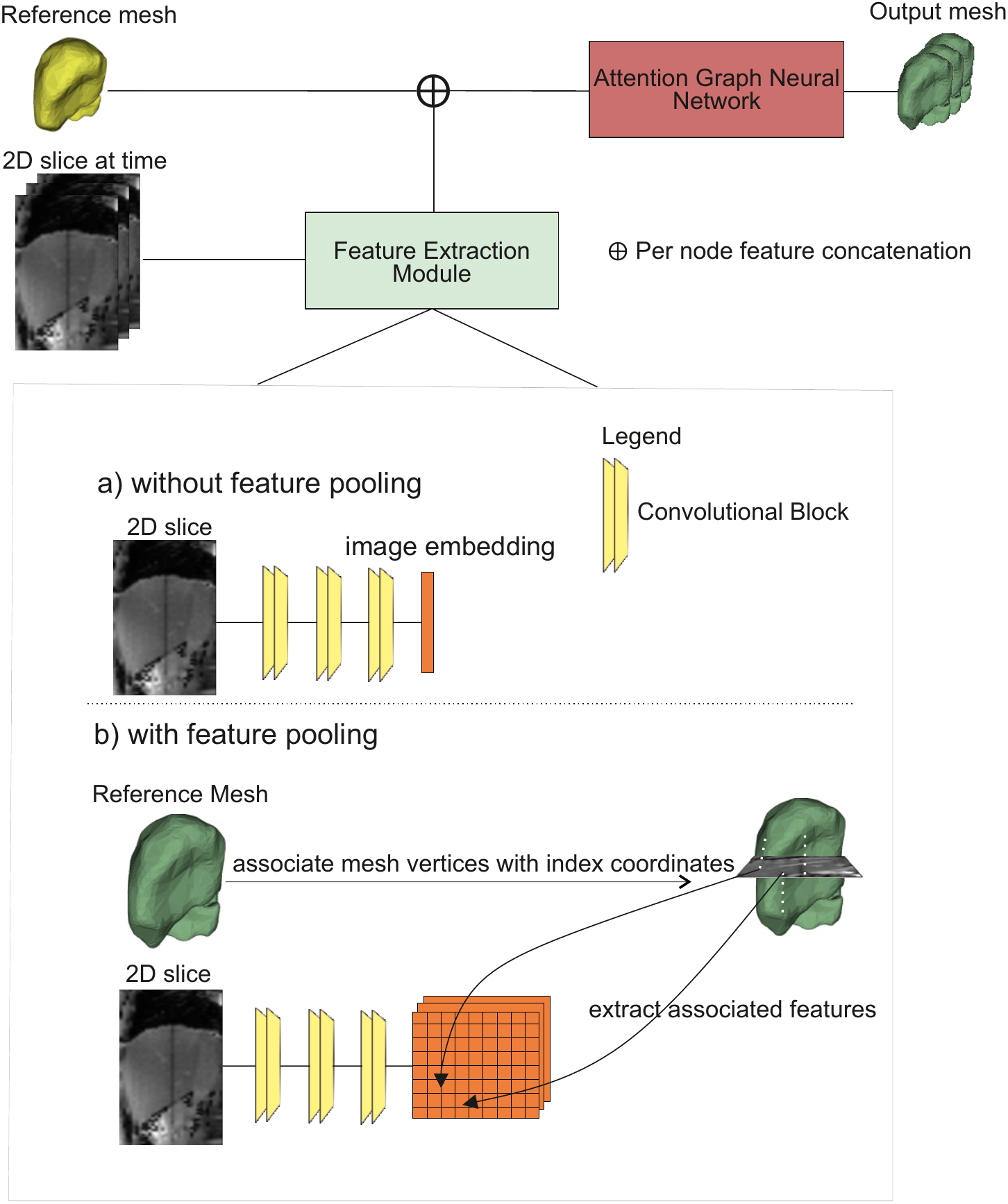}
    \caption{High-level representation of the DAGNN-SVLR model. The Attention Graph Neural Network deforms the reference mesh from the preoperative phase based on features extracted from supplied 2D surrogate images.
     We proposed two approaches to feature extraction: without feature pooling and with feature pooling.}
    \label{fig:high_level_model}
\end{figure}

In this work, we propose a Deformable Attention Graph Neural Network - Single View Liver Reconstruction (DAGNN-SVLR) method, which is an end-to-end trainable model that infers a 3D liver mesh structure at any time during treatment, using a reference triangulated mesh obtained from the segmentation of baseline 3D MRI volume and a single 2D MRI slice captured in real-time. Moreover, we present a new identity loss tailored for this specific task and empirically show that the combination with the Chamfer distance favors good mesh properties and improves the performance of the predictive model. A high-level representation of the proposed approach is shown in Fig. \ref{fig:high_level_model}.

\section{DAGNN-SVLR} 

The DAGNN-SVLR model leverages a triangulated liver mesh segmented from the pre-operative T2-w MRI volume and a 2D cine-MR slice taken in real-time during treatment. The model learns a function $f(.)$ that predicts the deformation of a reference mesh $M_r$ based on the surrogate 2D MRI slice at time $t$, defined as $I_t$, calculating the mesh at time $t$ as $M_t = f(M_r, I_t)$. 


A major component of the DAGNN-SVLR model is to determine correspondences between the reference mesh and a surrogate image. Since GNNs work directly on graphs and we cannot simply merge the image with the graph, we explored two solutions as illustrated in Fig. \ref{fig:high_level_model}.

\begin{enumerate}[(a)]
\item As a first option - without the feature pooling, we utilize a residual convolutional network as a feature extractor. The output is a latent representation of the image represented by a one-dimensional vector of size $dim=128$. ResNet18 \cite{he2016deep} network was selected as a compromise between accuracy and speed. 

\item As a second option, we propose a feature pooling approach. We use a ResNet18 network, with added padding, so each consequent layer of the network does not downsample the shape. Consequently, the feature maps produced by this ResNet have an identical shape to the input image. In parallel to the feature extraction, the index coordinates of the reference mesh are calculated, so that each vertex of the mesh can be associated with its particular position in the image. Afterward,  direct $3\times3$ neighborhoods are extracted from each feature map, thereby yielding nine features per feature map for each node in the reference mesh.
\end{enumerate}

Once the feature extraction module is processed, the concatenated features of the reference mesh, with the extracted features from the image, are passed through an attention graph neural network to produce the predicted mesh surface.

\subsection{Graph Convolutional Neural Network}
A Graph Convolutional Neural Network (GCNN) is a multilayer neural network that operates directly on graphs. It induces vertex embeddings based on the properties of their local neighborhood and the vertices.

Let $G=(V, E)$ be a triangulated surface of the liver volume, where $V = \{1, 2, \dots, N\}$  and $E \subseteq |V| \times |V|$ represent the set of vertices and edges, respectively. Let $X \in \mathbb{R}^{N x m}$ be a matrix containing the $m$ features of all $N$ vertices. We denote $\mathcal{N}_i = \{j: (i, j) \in E\} \cup \{i\}$ as the neighbor set of vertex $i$. Then, $H = \{\vec{h}_1, \vec{h}_2, ..., \vec{h}_N\}$ is a set of input vertex features, where feature vector $\vec{h}_i $ is associated with a vertex $i \in V$. 



Kipf et al. \cite{kipf2016semi} proposed a Spatial Graph Convolution layer GCN, where the features from neighboring vertices are aggregated with fixed weights, where one vertex embedding is calculated as:
\begin{equation}
    h_u = \Phi(\vec{x}_u, \bigoplus_{v \in \mathcal{N}_u} c_{uv} \psi (\vec{x}_v))
\end{equation}
where $\Phi$ and $\psi$ are learnable functions, $\vec{x}_u$ and $\vec{x}_v$  are vertex representations of vertex $u$ and vertex $v$, $c_{uv}$ specifies the importance of vertex $v$ to vertex $u$'s representation, $\mathcal{N}_u$ is a local neighborhood and $\bigoplus$ is an aggregation function, such as a summation or mean. 

Graph Attention Layers (GAT) \cite{velivckovic2017graph} extend this work by incorporating a self-attention mechanism that computes the importance coefficients $a_{uv}$: 
\begin{equation}
    h_u = \Phi(\vec{x}_u, \bigoplus_{v \in \mathcal{N}_u} a(\vec{x}_u, \vec{x_v}\psi(\vec{x}_v))).
\end{equation}

The attention mechanism $a$ is a single-layer feed-forward neural network parameterized by a weight vector $\vec{a} \in \mathbb{R}^{2F'}$, where:

\begin{equation}
    a_{ij} = \exp(LeakyRELU(\vec{a}^T)).
\end{equation}

We employ a neural network consisting of seven alternating GAT and normalization layers. Each GAT layer receives as input $128$ features and two heads, which output is summed afterward. According to the proposal in \cite{corso2020principal} we aggregate the features based on a combination of two aggregation functions: summation and mean.

\subsection{Loss functions}
We define three different loss functions to constrain the properties of output liver meshes. Amongst the most critical properties of final meshes are smoothness, closeness, and the absence of so-called flying vertices. 

\subsubsection*{Chamfer distance}
The Chamfer distance measures the distance of each vertex to the other
set and it serves as a constraint for the location of mesh vertices. 
The function is continuous, piecewise smooth and is defined as:
\begin{equation}
    \mathcal{L}_{CD}(P,Q) = \sum_p \min_q ||p-q||_2^2 + \sum_q \min_p ||p-q||_2^2.
\end{equation}
where $P$ and $Q$ are predicted and ground truth liver meshes and $p$ and $q$ are single points. Contrary to its name, based on the mathematical formulation, it is not a distance function since it does not hold the property of triangle inequality. 

\subsubsection*{Sampled Chamfer Distance} 
The Chamfer distance, despite its wide usage in the mesh processing domain, is unable to capture any fine information included in the mesh structure. It penalizes the point difference directly resulting in the loss of surface mesh information. 

To mitigate this drawback, Smith et al. \cite{smith2019geometrics} introduced a training objective that operates on a local surface defined by vertices sampled by a differentiable sampling procedure. 

Given a facet defined by 3 vertices $\{v_1, v_2, v_3\} \in \mathbb{R}^3$, uniform sampling is achieved by:

\begin{equation}
s = (1 - \sqrt{r_1})v_1 + (1-r_2)\sqrt{r_1} v_2 + \sqrt{r_1} r_2 v_3
\end{equation} where s is a point inside the surface defined by the facet, $r_1, r_2 \sim U[0,1]$

The loss function is defined as:
\begin{equation}
    \mathcal{L}_{SCD}(P,Q) = \sum_{p \in S} \min_{\hat{f} \in P} dist(p, \hat{f}) + \sum_{q \in \hat{S}} \min_{f \in Q} dist(q, f)
\end{equation}
where $P$ is the first mesh, $Q$ is the ground truth mesh, $\hat{S}$ and $S$ are the sampled points of the predicted mesh and ground truth mesh $f$ and $\hat{f}$ are faces and $dist$ is a function computing the distance between a point and a triangular facet.

\subsubsection*{Identity loss}
The identity loss penalizes substantial changes in the vertex positions if the surrogate image represents the actual state of the mesh. Given a mesh $M_t$, the surrogate signal at time $t$ $I_t$, and model that infers current mesh $\hat{M}_t$ based on surrogate image and a reference mesh $M_r$, $\hat{M}_t = f(M_r, I_t)$ we define identity loss as: 

\begin{equation}
\mathcal{L}_I = \mathcal{L}_{CD}(M, \hat{M})
\end{equation}
where $L_{CD}$ is a Chamfer loss or a sampled Chamfer loss. 

Finally, DAGNN calculates the final loss as $\mathcal{L} = \mathcal{L_{SCD}} + \alpha \mathcal{L}_I$ where $\alpha$ is a hyperparameter. 




\section{Results}
\label{sec:format}
We evaluated the proposed approach using a 4D-MRI liver dataset acquired from $25$ volunteers \cite{romaguera2021probabilistic}. The volume dimensions were $176 \times 176 \times 32$ with a pixel spacing of $1.7 \times 1.7 \textrm{mm}^2$ and a slice thickness of $3.5$ mm. Reference meshes were created as a closed surface of the liver segmentation from each of the subject's inhale phases. Ground truth meshes of other temporal sequences were then acquired by using deformation field calculated by Elastix deformable registration \cite{jansen2019liver}. Volumes were resized for the registration to $64 \times 64 \times 32$ due to computational complexity. The number of vertices in meshes were from interval $(1300, 2000)$. When sampling loss was used, an empirically determined value of $1000$ was chosen. 

As for preprocessing steps, we centered and normalized the scale into the interval $(-1, 1)$. For validation purposes, we performed a 10-fold cross-validation. The model was trained with a batch size of one with five steps accumulation gradients and with Adam optimizer with a learning rate $1e^{-5}$. We set the weight of identity function $L_I$ to $0.05$. 

\begin{figure}[h]
    \centering
    \begin{subfigure}[t]{0.41\textwidth}
        \hspace{3mm}\raisebox{-\height}{\includegraphics[width=0.28\textwidth]{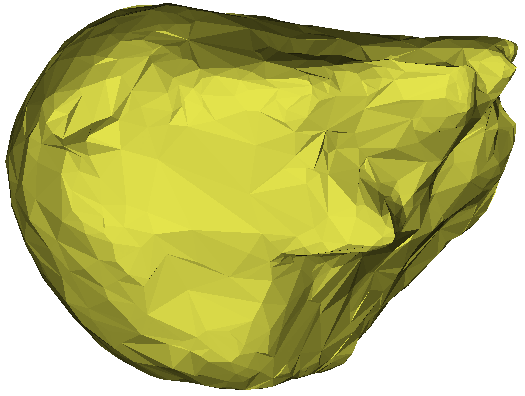}}
        \hspace{3mm}
        \raisebox{-\height}{\includegraphics[width=0.28\textwidth]{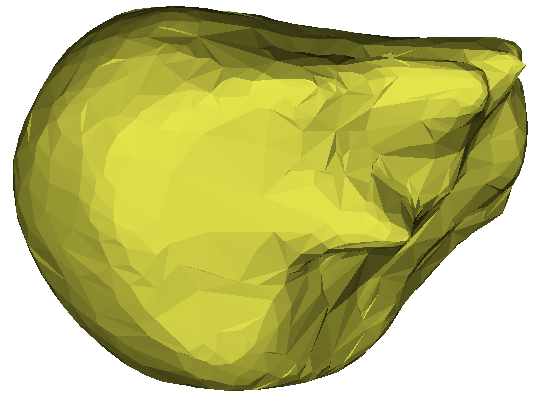}}%
        \hspace{3mm}
        \raisebox{-\height}{\includegraphics[width=0.28\textwidth]{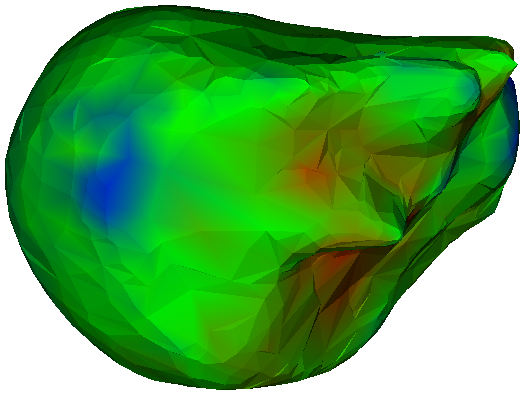}}%
        \vspace{2mm}
        \\
        \hspace{10mm}
        \raisebox{-\height}
        {\hspace{4mm}\includegraphics[width=0.25\textwidth]{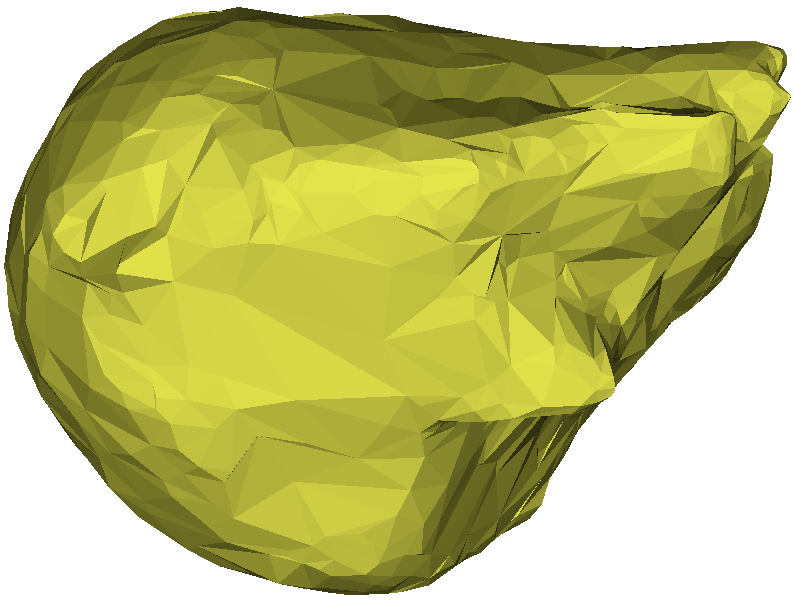}}
        \hspace{4mm}
        \raisebox{-\height}{\includegraphics[width=0.25\textwidth]{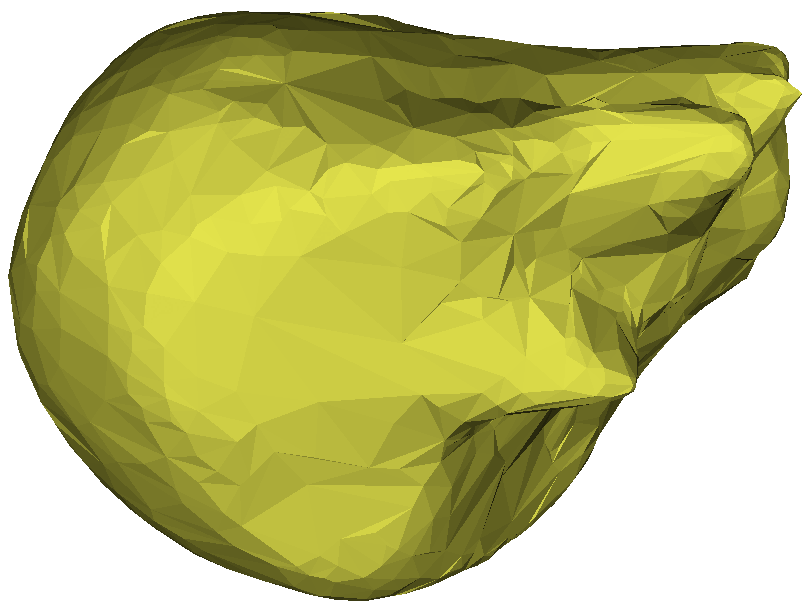}}
        \hspace{3mm}
        \raisebox{-\height}{\includegraphics[width=0.25\textwidth]{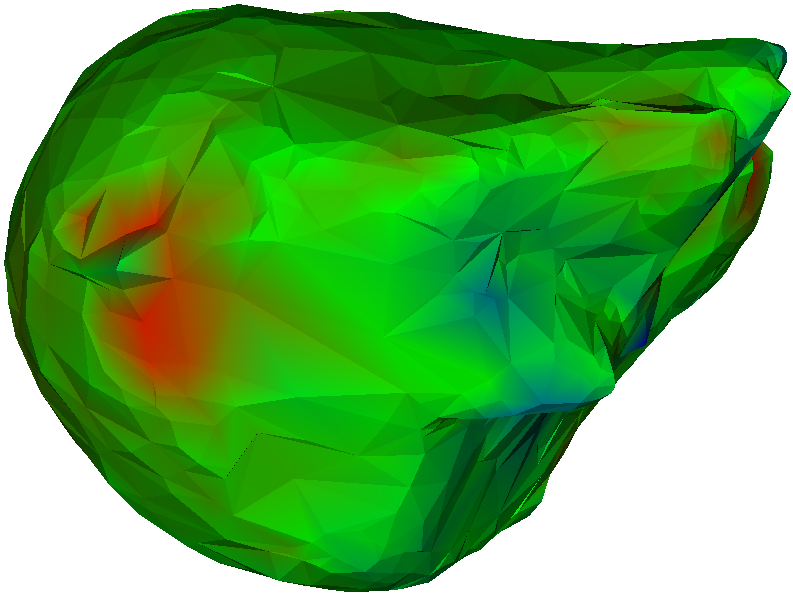}}
        \begin{tabular}{p{2cm}p{2.56cm}p{1.4cm}}
         \vspace{2mm}\centering Ground Truth \hfill & \vspace{2mm} \centering Predicted Mesh \hfill & \centering \vspace{0.3mm} Signed Distance \hfill \\
         \end{tabular}
    \end{subfigure}
    \hfill
    \begin{subfigure}[t]{0.05\textwidth}
        \vspace{0mm}\raisebox{-\height}{\includegraphics[width=0.85\textwidth]{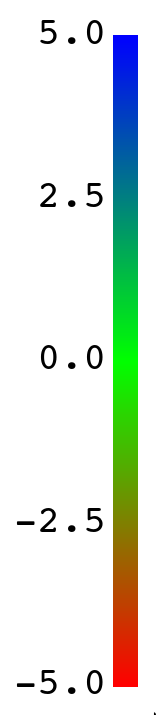}}
    \end{subfigure}
    
    \caption{Visualization of a signed distance (in mm) from predicted to ground mesh for two subjects.}
    \label{viz}
\end{figure}

\begin{figure}[h!]%
        \includegraphics[height=0.15\textheight, width=0.12\textwidth]{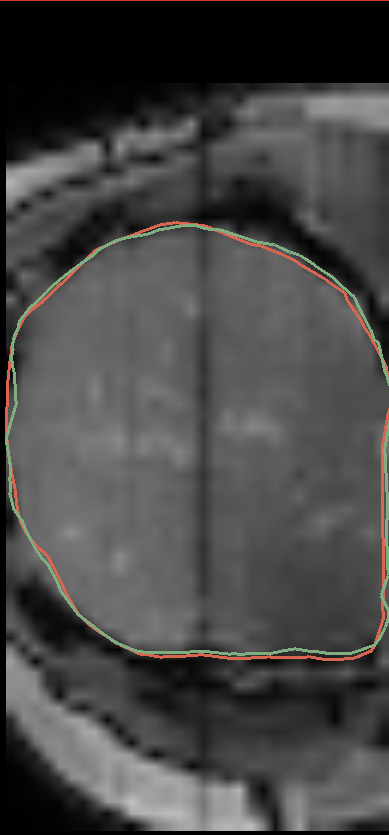}
        \includegraphics[height=0.15\textheight, width=0.12\textwidth]{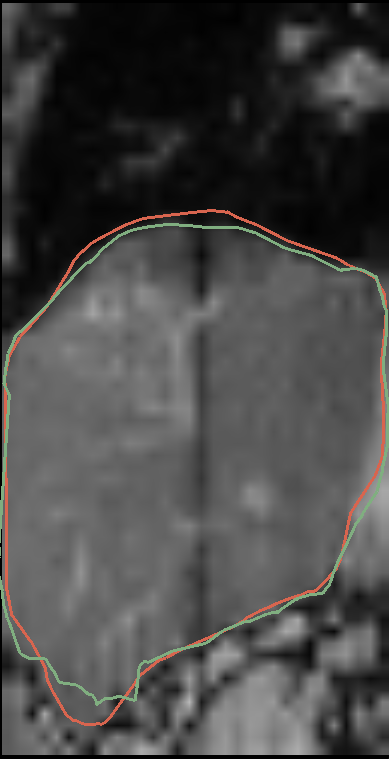} 
        \includegraphics[height=0.15\textheight, width=0.23\textwidth]{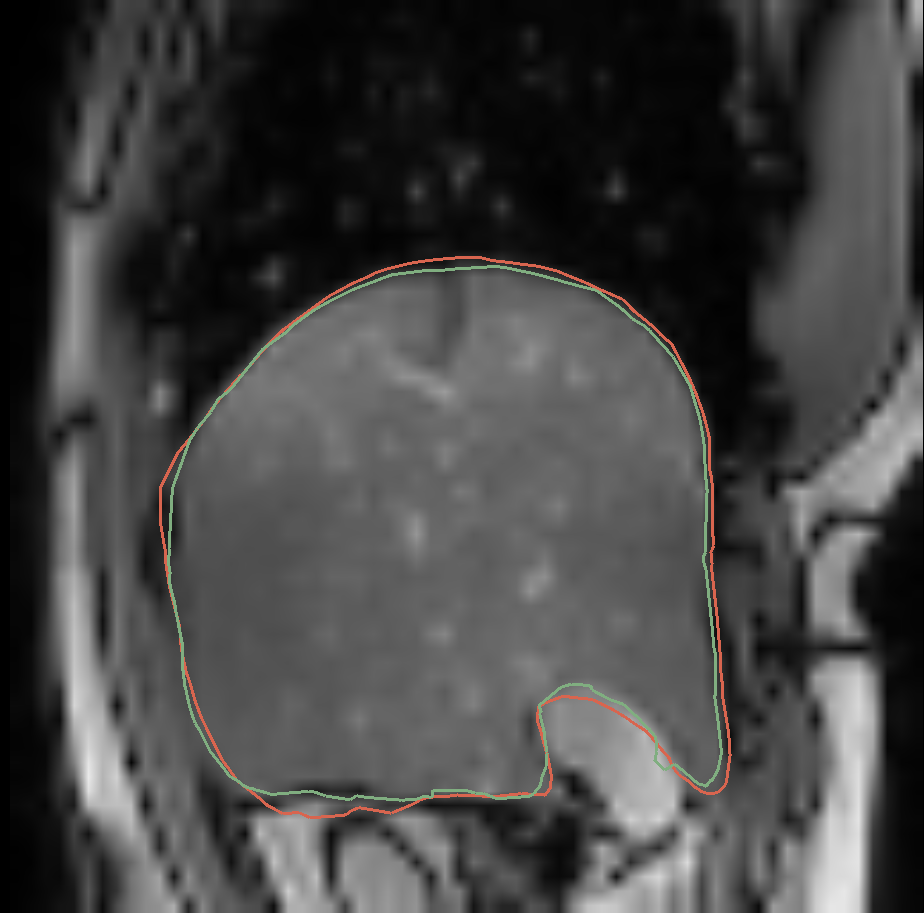} \\

        \includegraphics[height=0.15\textheight, width=0.12\textwidth]{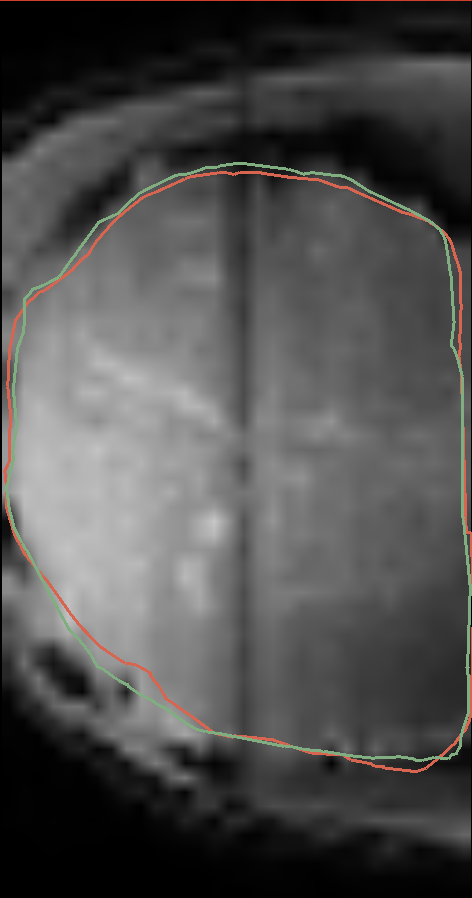}
        \includegraphics[height=0.15\textheight, width=0.12\textwidth]{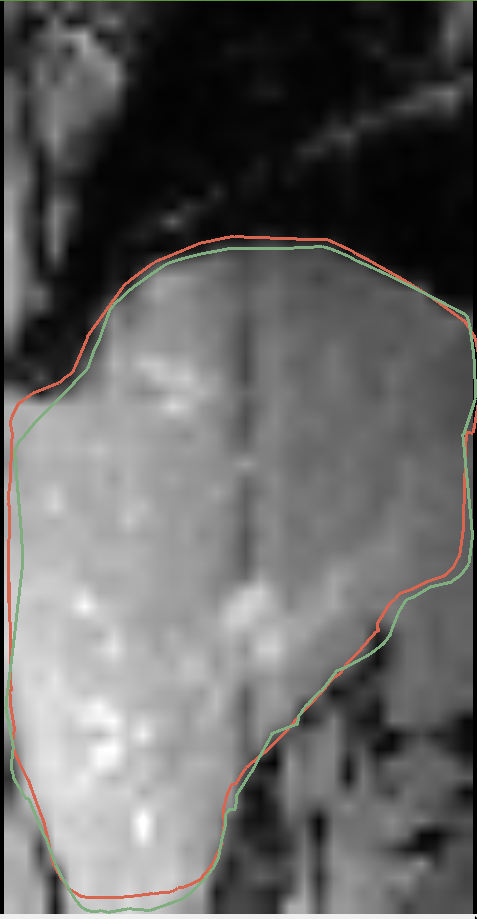} 
        \includegraphics[height=0.15\textheight, width=0.23\textwidth]{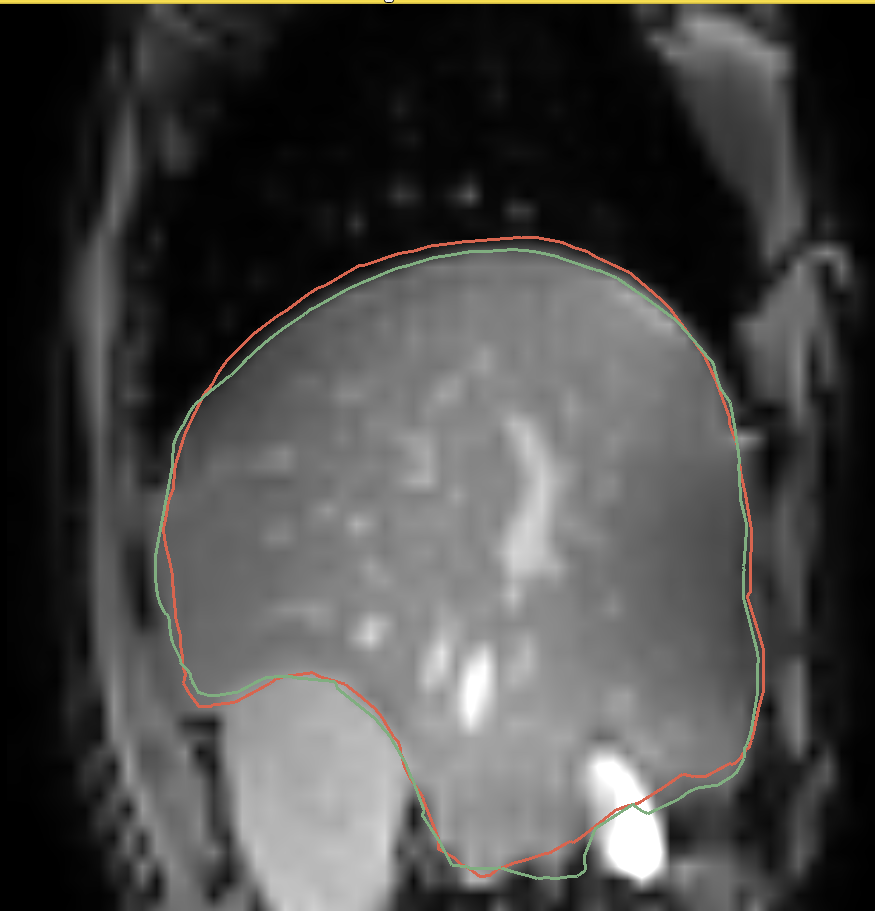}

        \caption{Comparison of ground truth (green) and predicted (red) segmentations. Each row depicts different volunteer acquisitions.}
        \label{viz2}
\end{figure}

Visualizations of sample predictions and their ground truth can be seen in Fig. \ref{viz}. It is clear that the inferred shapes are compliant with important mesh properties such as complete and smooth surfaces. Additionally, in Fig. \ref{viz2} we show ground truth and predicted delineations obtained from inferred mesh in the axial, sagittal, and coronal planes.

The average error and average Chamfer distance over the entire test sequence is presented in Table \ref{table:results}. We used the Chamfer distance with L2 norm and errors calculated by unsigned distance of two meshes to measure the performance. The results of our model are divided into two main categories, with and without feature pooling.

\begin{table}[h!]
\centering
\resizebox{\columnwidth}{!}{
\begin{tabular}{|c| c c|} 
 \hline
 \textbf{Method} & \textbf{Chamfer distance with L2 norm} & \textbf{Avg. error (mm)} \\ [0.5ex] 
 \hline
  Node2Vec \cite{grover2016node2vec} & $79.45 \pm 48.67$ & $5.0 \pm 3.7$\\

 FP + $\mathcal{L}_{CD}$ & $62.85 \pm 25.18$ & $3.44 \pm 0.91 $ \\
 FP + $\mathcal{L}_{SCD}$  & $82.25 \pm 24.40$ & $3.23 \pm 0.64$ \\
 FP + $\mathcal{L}_{SCD}$ + $\mathcal{L}_I$ & $63.14 \pm 27.28$ & $\mathbf{3.05 \pm 0.75}$ \\
 No FP + $\mathcal{L}_{CD}$ & $67.226 \pm 5.57$ & $3.46\pm0.70$  \\ [1ex] 
 No FP + $\mathcal{L}_{SCD}$ & $66.117 \pm 26.48$ & $3.26 \pm 0.78$  \\ [1ex] 
 No FP + $\mathcal{L}_{SCD}$ + $\mathcal{L}_I$ & $\mathbf{61.34 \pm 25.63} $ & $3.06 \pm 0.7$  \\ [1ex] 
 \hline
\end{tabular}
}

\caption{Prediction results for different loss functions, with and without feature pooling (FP).}
\label{table:results}
\end{table}

To compare our method with a state-of-the-art approach, we chose the Node2Vec \cite{grover2016node2vec} method that supports working with graph data. Node2Vec was trained for $100$ epochs and found embeddings were concatenated with extracted features, similarly to our model.

Feature pooling, as used in this study, slightly outperforms the utilization of a single feature vector with respect to average error in all cases. This confirms our hypothesis, that leveraging sampling loss improves performance for both feature extraction methods. The average error for feature pooling subsided from $3.44 \pm 0.91$mm to $3.23 \pm 0.64$mm, and for the approach without FP from $3.46 \pm 0.70$mm to $3.26 \pm 0.78$mm. Adding identity loss during the training improves the results from $3.23 \pm 0.91$mm and $3.26 \pm 0.78$mm to $3.05 \pm 0.75$mm and $3.06 \pm 0.7$mm respectively, thus representing a performance improvement of $5.6$\% and $6.2$\%.

Interestingly, the test results of Chamfer distance are higher when the sampling loss was used without identity loss for feature pooling. We hypothesize that the model was trained using a loss that sampled points on the surface, but for the testing part conventional Chamfer loss was used. This error in Chamfer distance diminishes when identity loss is added.

\begin{figure}[h]
    \centering
    \includegraphics[width=0.49\textwidth]{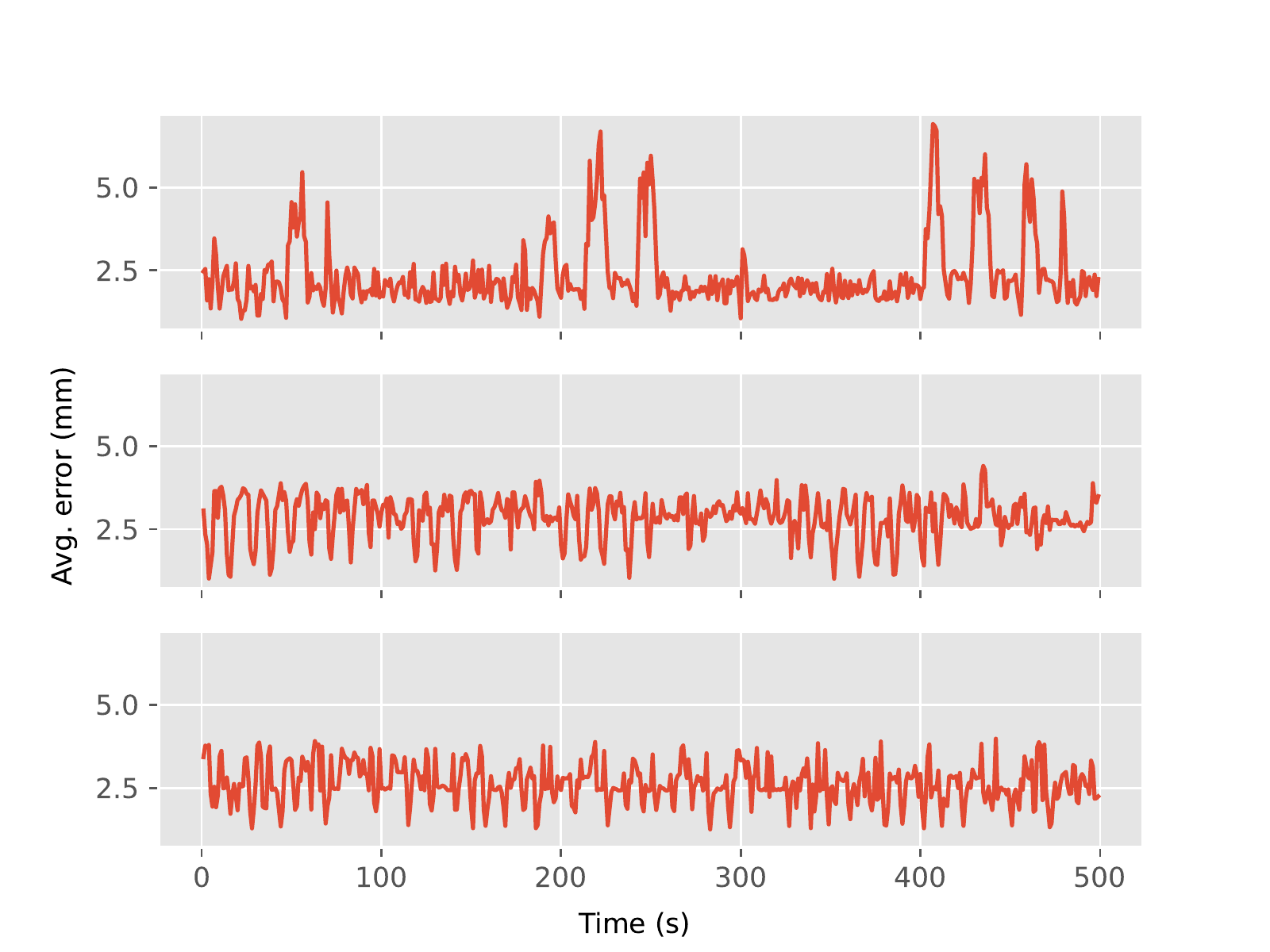}
    \caption{Average error (in mm) over free-breathing sequences for three selected subjects.}
    \label{fig:test}
\end{figure}

Next, we present the average error in millimeters for three subjects over time in Fig. \ref{fig:test}. The plot of the subject in first row has spikes that introduce errors yielding more than $5$ mm.
The other two subjects have very similar errors over time oscillating over a value of $2.5$mm without any outliers. The model repeatedly achieve error as low as $1.8$mm.

\section{Conclusion}
We presented a novel approach for single-view liver surface reconstruction from a surrogate 2D signal based on the combination of an attention graph neural network with a fully convolutional neural network. Our model was successful in generating full meshes with proper topology, and position and with an inference time of 0.002 seconds. We have shown that the model benefits considerably from the proposed identity loss, with a feature pooling process. Several potential extensions will be addressed in future work, such as liver motion prediction in the graph domain or using a sequence of surrogate images for liver reconstruction, instead of using a single view.



\section{Compliance with Ethical Standards}
This study was performed in line with the principles of the Declaration of Helsinki. Approval was granted by the local Institutional Review Board.

\section{Acknowledgments}
\label{sec:acknowledgments}

This research has been funded in part by the Natural Sciences and Engineering Research Council of Canada (NSERC). The authors have no relevant financial or non-financial interests to disclose.

\bibliographystyle{IEEEbib}
\bibliography{strings,refs}

\end{document}